%% file: paper.tex
\documentclass[conference]{IEEEtran}
\IEEEoverridecommandlockouts
\usepackage{caption}
\usepackage{subcaption} 
\usepackage{cite}
\usepackage{url}
\usepackage[breaklinks,colorlinks=true, allcolors=blue]{hyperref}
\usepackage{amsmath,amssymb,amsfonts}
\usepackage{algorithmic}
\usepackage{graphicx}
\usepackage{textcomp}
\usepackage{lipsum} 
\usepackage{xcolor}
\usepackage{float}
\usepackage{booktabs} 
\usepackage{mdframed}
\usepackage{multicol} 
\usepackage{subcaption}
\usepackage{stfloats}
\usepackage[ruled,vlined]{algorithm2e}

\def\BibTeX{{\rm B\kern-.05em{\sc i\kern-.025em b}\kern-.08em
    T\kern-.1667em\lower.7ex\hbox{E}\kern-.125emX}}

\makeatletter 
\newcommand{\linebreakand}{
  \end{@IEEEauthorhalign} 
  \hfill\mbox{}\vspace{3mm}\par
  \mbox{}\hfill\begin{@IEEEauthorhalign}
}
\makeatother 

\begin{document}

\title{Improving VTE Identification through Language Models from Radiology Reports: A Comparative Study of Mamba, Phi-3 Mini, and BERT \thanks{* Corresponding author}
 }

\author{\IEEEauthorblockN{Jamie Deng$^\dagger$, Yusen Wu$^{\ddagger,*}$, Yelena Yesha$^{\dagger,\ddagger,\S}$, Phuong Nguyen$^{\dagger,\ddagger}$}
\IEEEauthorblockA{$\dagger$\textit{Department of Computer Science}, University of Miami, Miami, FL USA\\
$\ddagger$\textit{Frost Institute for Data Science \& Computing}, University of Miami, Miami, FL, USA\\
$\S$\textit{Department of Radiology}, University of Miami, Miami, FL, USA\\
\{jxd3987, yxw1259, yxy806, pnx208\}@miami.edu}}

\IEEEoverridecommandlockouts
\maketitle
\IEEEpubidadjcol

\begin{abstract}

Venous thromboembolism (VTE) is a critical cardiovascular condition, encompassing deep vein thrombosis (DVT) and pulmonary embolism (PE). Accurate and timely identification of VTE is essential for effective medical care. This study builds upon our previous work, which addressed VTE detection using deep learning methods for DVT and a hybrid approach combining deep learning and rule-based classification for PE. Our earlier approaches, while effective, had two major limitations: they were complex and required expert involvement for feature engineering of the rule set. To overcome these challenges, we utilize the Mamba architecture-based classifier. This model achieves remarkable results, with a 97\% accuracy and F1 score on the DVT dataset and a 98\% accuracy and F1 score on the PE dataset. In contrast to the previous hybrid method on PE identification, the Mamba classifier eliminates the need for hand-engineered rules, significantly reducing model complexity while maintaining comparable performance. Additionally, we evaluated a lightweight Large Language Model (LLM), Phi-3 Mini, in detecting VTE. While this model delivers competitive results, outperforming the baseline BERT models, it proves to be computationally intensive due to its larger parameter set. Our evaluation shows that the Mamba-based model demonstrates superior performance and efficiency in VTE identification, offering an effective solution to the limitations of previous approaches.
\end{abstract}


\input{intro}
\input{method}

\input{results}

\input{related}
\input{future}
\input{conclusion}

\section{Acknowledgement}
We would like to express our gratitude to the medical experts from the University of Maryland Medical Center (UMD) for their invaluable assistance in collecting and annotating the VTE datasets used in this study. Their expertise and dedication have been instrumental in the development and validation of our research. Thanks to the Department of Radiology at University of Miami for internal funding support.

\bibliographystyle{ieeetr}
\bibliography{refs}

\end{document}

%% file: intro.tex

\section{Introduction}

Venous thromboembolism (VTE) \cite{cohen2007venous}, encompassing deep vein thrombosis (DVT) and pulmonary embolism (PE), is the third most common cardiovascular condition globally \cite{nelson2015using}. DVT is characterized by the formation of a blood clot within a deep vein, commonly affecting the lower extremities, while PE occurs when a clot detaches and travels to the lungs through the bloodstream. VTE significantly complicates surgical procedures and leads to longer hospital stays and higher mortality rates if undetected \cite{woller2021natural}. The risk of VTE can increase up to 20-fold after surgery \cite{white2002risk}. Therefore, the prompt identification of VTE is crucial for medical decision-making, and the adoption of automated methods for diagnosing VTE could significantly enhance healthcare practices.


The wide adoption of electronic health record systems (EHRs) across US hospitals offers a significant opportunity to utilize advanced data analytics for the classification of postoperative VTE. Clinical notes and reports contain vital details about postoperative complications \cite{shi2021natural}. To extract valuable information from these unstructured and free-text documents, Natural Language Processing (NLP) employs computational linguistics to analyze the textual data. The use of NLP has been increasingly common in the analysis of radiologist reports from medical imaging \cite{pons2016natural}. Given that the diagnosis of VTE heavily relies on imaging findings, the application of NLP can help in automatically identifying patients with VTE through radiology reports.

In our previous study \cite{deng2023improving} of VTE identification by fine-tuning deep learning models, we developed a system that employs ClinicalBERT \cite{alsentzer2019publicly} and Bi-LSTM \cite{graves2005framewise} to identify DVT, and integrates these deep learning models with a rule-based classifier (hybrid model) to detect PE from unstructured free-text radiology reports. Our results demonstrate the system's effectiveness, achieving high accuracy and F1 scores. However, the work is still facing some challenges. The system is complex with multiple components. The PE rule set requires manual feature engineering by clinical experts, which restricts its generalization ability when applied to other medical domains. Furthermore, the input sequence length of BERT models is limited to 512 tokens, posing challenges when applying the model to longer textual data. Any medical reports exceeding this length are truncated, potentially losing important contextual information.

To tackle these challenges, we aim to:
\begin{itemize}
    \item Reduce Components for Efficiency: This involves simplifying the classifier's architecture to reduce computational complexity and enhance processing speed.
    \item Expand the Context Window of the Model: Enhancing the model's capacity to consider a broader range of contextual information can significantly improve its performance and accuracy in understanding complex patterns and relationships within the data.
    \item Eliminate Rule Sets for Generalization: By removing rigid rule-based classifier, the model can adapt more flexibly to various scenarios, improving its ability to generalize across different contexts and data types. 
\end{itemize}

To achieve these objectives, a new architecture for the classifier is essential. This new design will focus on integrating these improvements seamlessly, ensuring that the classifier not only performs effectively but also adapts to different situations.

Transformers \cite{vaswani2017attention}, popular in advanced Language Models like BERT and LLMs, have limitations despite their innovative self-attention mechanism. These include higher computational costs and longer training times, especially with big data or complex tasks. They also struggle to model relationships beyond a certain range, with complexity growing rapidly as the length of input sequence increases. Recently, novel architecture called Mamba \cite{gu2023mamba}, inspired by State Space Models (SSM), offers improved inference speed and scalability for long sequences due to its distinctive selection mechanism. This mechanism enables Mamba to perform context reasoning and selectively focus on specific inputs, thereby lowering computational costs and boosting performance \cite{song2024state}. It also makes Mamba particularly effective for complex tasks and handling long sequence datasets. However, its application in medical text classification remains largely unexplored. Mamba is a promising alternative to Transformers for creating foundation models, providing comparable modeling capabilities while maintaining near-linear scalability in both training and inference \cite{qu2024survey}.

In this paper, we employ a Mamba-based model for the VTE classification tasks. We use the pre-trained Mamba-130M model as the base and add a linear layer on top as classification head. Then we fine-tune the model on the VTE datasets. This approach streamlines the training process compared to our previous work \cite{deng2023improving} and extends the context window to accommodate longer radiology reports, thanks to Mamba's extensive sequence length of 8K tokens. The proposed model produces comparable performance to the hybrid model from our previous work. Therefore it removes the need for a rule based classifier which requires an expert-selected rule set. 

We also investigate the classification capabilities of a smaller LLM, Phi-3 Mini \cite{abdin2024phi}, which has 3.8 billion parameters. For comparison, we include two Transformer-based classifiers, DistilBERT \cite{sanh2019distilbert} and DeBERTa \cite{he2021debertav3}, as baseline models.

The Mamba classifier demonstrated excellent performance with a 97\% accuracy and F1 score on the DVT dataset and a 98\% accuracy and F1 score on the PE dataset. The Phi-3 Mini classifier also outperformed the BERT models. However, its larger number of parameters makes it computationally demanding. For text classification tasks, using a Large Language Model is not the most efficient approach. The experimental results support the efficacy of Mamba-based models for medical text classification tasks.

We summarize our contributions as follows:
\begin{itemize}
  \item Our research addresses the complexity of previous approaches by developing a more streamlined model architecture. The Mamba-based classifier significantly reduces the overall complexity compared to prior methods, which relied on hybrid models combining deep learning with rule-based systems.

\item  We eliminate the need for expert-engineered rule sets that were essential in previous methods. This generalizes the model's applicability across various domains and reduces the dependence on domain experts for manual feature engineering.

\item  We conducted extensive experiments demonstrating the superior performance of the Mamba-based classifier. The model outperforms all previous methods in terms of accuracy and F1 score while also proving to be more efficient than LLMs, making it an optimal solution for VTE classification tasks.
    
\end{itemize}

%% file: method.tex
\section{Proposed Methods}

We propose to utilize the Mamba \cite{gu2023mamba} architecture based classifier to do text classification on the VTE datasets of radiology reports. The Transformer architecture has played a crucial role in the success of Language Models and has widely applied in many different NLP tasks, powering nearly all widely used models today, such as LLM and BERT. To push the boundaries of Language Models, researchers are exploring new architectures that could surpass the Transformer. One promising approach is Mamba architecture.

\subsection{SSM Fundamentals}
Mamba is a State Space Model (SSM) architecture that demonstrates promising performance on information-rich tasks like language modeling, where earlier sub-quadratic models have struggled to match the effectiveness of Transformers. It builds on advancements in structured state space models and features a hardware-optimized design and implementation, inspired by the efficiency of Flash Attention \cite{dao2022flashattention}. Mamba offers a more efficient approach for long sequences with its linear complexity and simplified architecture, making it a promising alternative for tasks requiring long-term dependencies such as textual data.

A SSM is a mathematical representation of a system using a set of input, output, and state variables. The state variables define the values of the output variables and evolve over time based on their current values and the input variables. A SSM is defined by these two equations. It maps a 1-D input signal $x(t)$ to an N-D latent state $h(t)$ before projecting to a 1-D output signal $y(t)$.
\begin{align*}
 h'(t) &= Ah(t) + Bx(t) \qquad\text{State equation}  \\
 y(t) &= Ch(t) + Dx(t)  \qquad\text{Output equation}
\end{align*}
SSM is used as a black-box representation in a deep sequence model, where A, B, C, D are parameters learned through gradient descent. They are in matrix format. Parameter D is usually omitted since it's similar to a skip connection. A SSM maps a input $x(t)$ to a state representation vector $h(t)$ and an output $y(t)$. Assume the input and output are one-dimensional, and the state representation is multi-dimensional. The state equation describes how $h(t)$ changes over time. The output equation defines how the state is translated to the output. By solving these equations, we can predict the state of a system based on observed data: input sequence and previous state. 

For language modelling, we need to use the discrete version of SSM and recurrent representation. SSM can also be represented as convolution therefore increasing training efficiency. The parameter A captures information about the previous state to build the new state. In order for matrix A to retains long history of states, HiPPO \cite{gu2020hippo} technique is used to address long range dependencies. It works by transforming input data into a higher-dimensional space using polynomial functions to capture complex relationships, which are then fed into neural networks for improved learning efficiency.

Mamba introduces two more innovations: Selective Scan Algorithm, which enables the model to focus on pertinent data by filtering out the irrelevant. Hardware-Aware Algorithm, which Enhances efficiency by streamlining storage and processing via parallel operations, kernel fusion, and recomputation. These innovations lead to the Selective SSMs (S6 models), which are utilized in Mamba blocks much like self-attention mechanisms in Transformers.

\subsection{Classification Model}
\begin{figure}[t] 
        \centering        \includegraphics[width=0.35\textwidth]{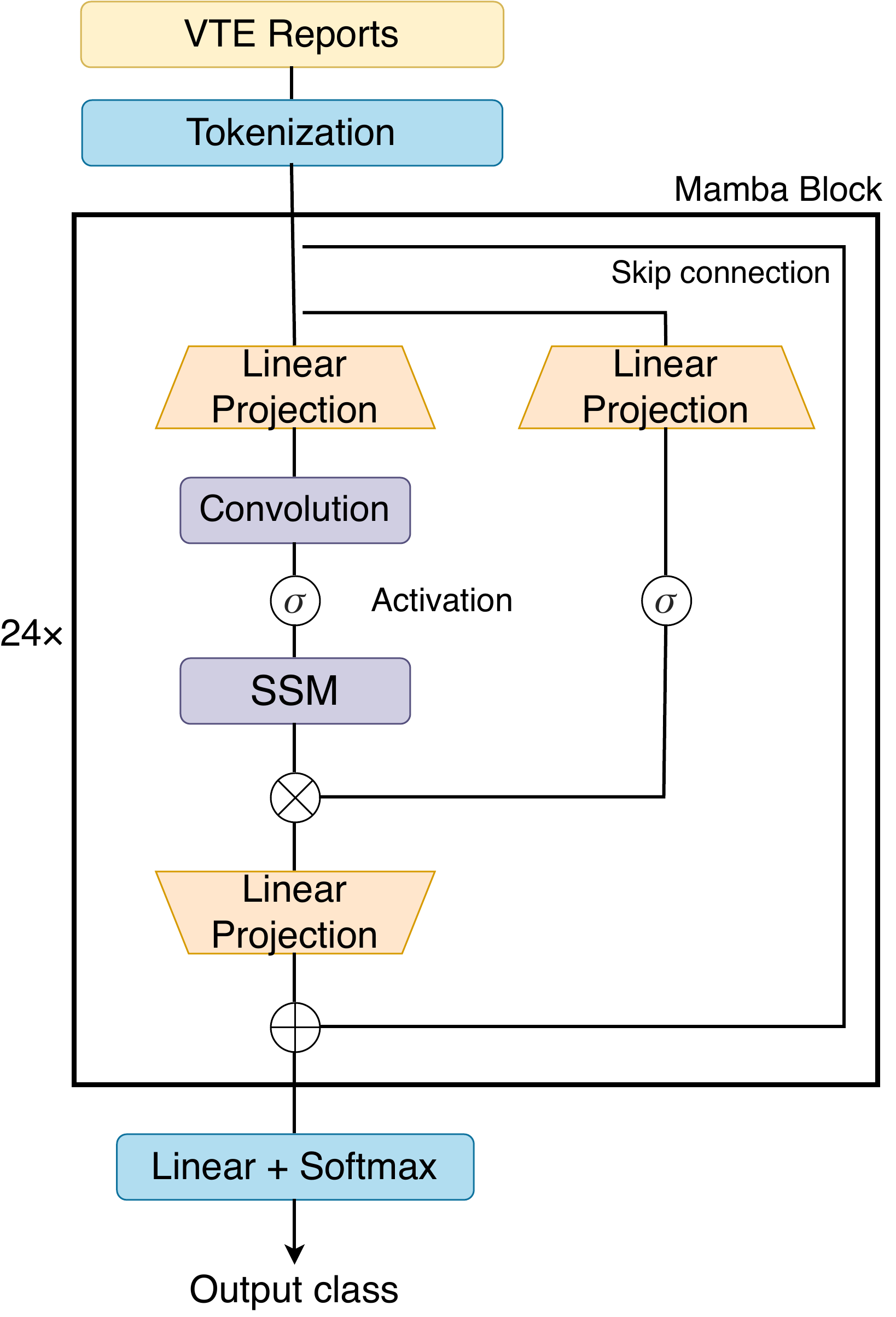}
        \caption{Structure of Mamba-based classifier. Mamba-130M contains 24 layers.}
        \label{fig:model}
    \end{figure}

Mamba architecture is particularly well-suited for handling long sequences of text, making it an ideal choice for the classification tasks of medical reports. For the backbone to the classifier, we chose the Mamba-130M model, which is the smallest version of Mamba with 130 million parameters. This is a significant reduction in size compared to our earlier research using ClinicalBERT and Bi-LSTM models. For instance, a standard BERT model has 110 million parameters, while a Bi-LSTM could have many more, depending on the input size and hidden layer dimensions. We add a linear layer to serve as the classification head, which allows the model to output predicted labels. Figure \ref{fig:model} demonstrate the structure of the classifier.

The Mamba block is a key component of the Mamba architecture, featuring linear projections that prepare the input sequence for further processing and a convolutional mode for efficient parallel training. At its core is the Selective SSM, which updates the sequence's state representation and focuses on key parts of the input. The SiLU (Sigmoid Linear Unit) activation introduces non-linearity after the convolution and SSM processing. The SSM output is combined with a gated projection output through multiplication, integrating different input aspects. Finally, the combined result undergoes a projection and adds a skip connection. The Mamba-130M model comprises 24 layers. Each block or layer processes the input sequentially, with the output of one block serving as the input for the next.
We fine-tune the entire model, including the pre-trained Mamba backbone and the added linear layer, on the radiology report datasets. This process allows the model to adapt to the specific characteristics of the datasets and learn to classify the text effectively. 

To explore the ability of Large Language Model (LLM) in the text classification tasks. We select one of the small LLMs for the NLP tasks. The Phi-3 Mini \cite{abdin2024phi}, with its 3.8 billion parameters, is lightweight, state-of-the-art open source model that can capture complex relationships and patterns in text data. By fine-tuning the Phi-3 Mini using the QLoRA \cite{dettmers2024qlora} method, we can adapt the model to our specific tasks. QLoRA (Quantized Low-Rank Adaptation) fine-tuning is a technique used particularly in the realm of LLM, to make the fine-tuning process more efficient and less resource-intensive. It enables the model to learn from a limited amount of labeled data, making it a practical choice for our research. To evaluate the effectiveness of the Mamba classifier, we compare it against two Transformer-based BERT models, DistilBERT \cite{sanh2019distilbert} and DeBERTa \cite{he2021debertav3}, as baselines for the classification tasks.


%% file: results.tex
\section{Experiment Results}

\subsection{VTE Datasets}
We use the same two datasets from our previous work \cite{deng2023improving}, which contain medical imaging reports for VTE classification (DVT and PE). These datasets comprise de-identified and labeled medical reports. They were sourced from the University of Maryland Medical Center (UMD). The de-identification and labeling of datasets were done by medical experts from UMD.

The first dataset includes 1,000 free-text duplex ultrasound imaging reports. The reports were classified into 3 categories by a Radiologist: Class 0 - No acute DVT, Class 1 - Upper extremity acute DVT, and Class 2 - Lower extremity acute DVT. A total of 78\% of data samples fall into the category of class 0, and 11\% for class 1 and 2 respectively. The dataset consists primarily of structured reports containing concise texts, with the majority of them being less than 170 words in length.

The second dataset includes 900 free-text chest computed tomography (CT) angiography scan reports. It has fewer samples than the first dataset and is more imbalanced. The reports were classified into 2 categories: class 0 - No PE (88\%), class 1 - PE (12\%).  These CT scan reports contain mostly unstructured texts and are longer in length. Most of them are around 200 words. Some reports exceed 600 words.

\subsection{Experimental Settings}
To assess the effectiveness of the Mamba classifier, we conducted two series of experiments. The first series employed the DVT dataset, composed of shorter, well-structured text from Ultrasound reports. In this series, we tested the Mamba-based classifier proposed in this study alongside several baseline classifiers. The second series of experiments focused on the PE dataset, which contains longer, more complex text from CT scan reports. This dataset is both limited in size and imbalanced.

We divided the datasets into 80\% training and 20\% test sets. The training sets were further split into 90\% training and 10\% validation sets. The input texts are truncated to match the input limits of the different classifiers. The DVT dataset consists of shorter texts, which are well within the input limit of all classifiers. However, the PE dataset contains longer texts that exceed the sequence length limit of BERT models. For BERT models, the maximum input sequence length is 512 tokens. In contrast, the Mamba model allows a maximum input length of 8,000 tokens, having been pre-trained on sequences of 2,000 tokens. As a result, the Mamba classifier can handle longer text sequences than BERT models.

\subsection{Model Performance}
We compare the proposed method with two baseline Transformer based BERT model as well as a small LLM. 

\input{table1}
\input{table2}

The baseline Transformer based classifiers include:
\begin{itemize}
    \item DistilBERT \cite{sanh2019distilbert}: A smaller pretrained general-purpose language representation model. It is a compact, efficient, and cost-effective Transformer based model, trained through the distillation of BERT base. It has 40\% fewer parameters than google-bert/bert-base-uncased, runs 60\% faster, and retains over 97\% of BERT’s performance.
    \item DeBERTa \cite{he2021debertav3}: It enhances the BERT and RoBERTa models by incorporating disentangled attention and an improved mask decoder. These advancements enable DeBERTa to outperform RoBERTa on most NLU tasks using 80GB of training data. Efficiency is further boosted through ELECTRA-style pre-training with Gradient Disentangled Embedding Sharing, leading to significant improvements in performance on downstream tasks.
    \item Phi-3 Mini 128k \cite{abdin2024phi}: It's a 3.8 billion-parameter, state-of-the-art, lightweight open model trained on the Phi-3 datasets. This dataset comprises both synthetic data and filtered publicly available website data, with a focus on high-quality, reasoning-intensive content. The model is part of the Phi-3 family, with the Mini version available in two variants, 4K and 128K, representing the context length (in tokens) it can handle.
\end{itemize}

\begin{figure*}[ht]
\centering
    \begin{subfigure}{.33\textwidth}
        \centering
        {\includegraphics[scale=0.38]{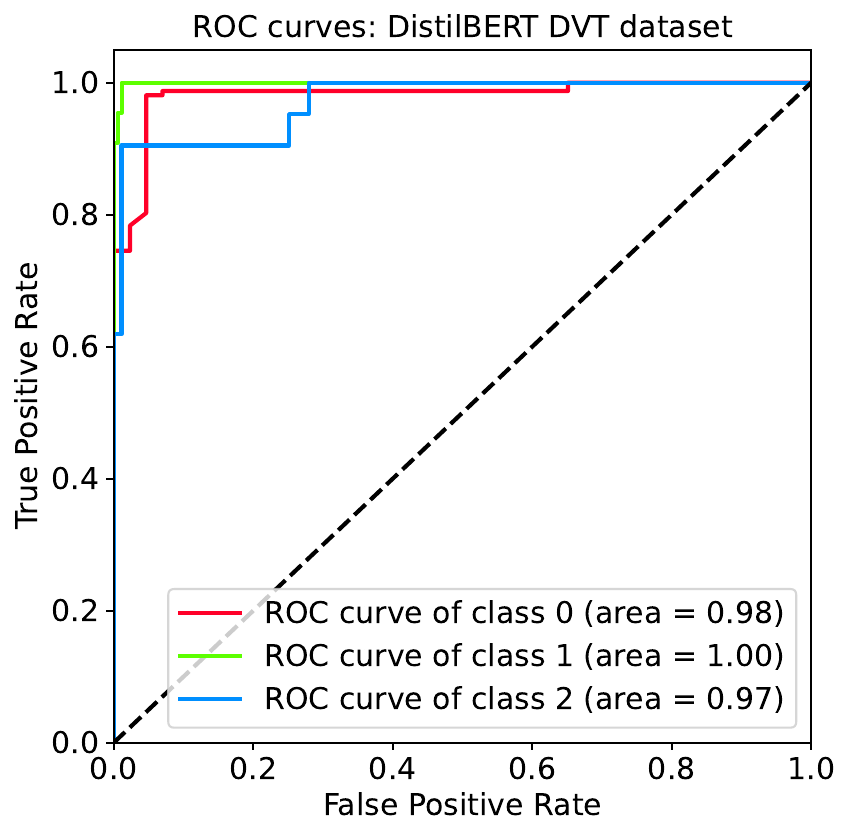}} 
    \end{subfigure}%
    \begin{subfigure}{.33\textwidth}
        \centering
        {\includegraphics[scale=0.38]{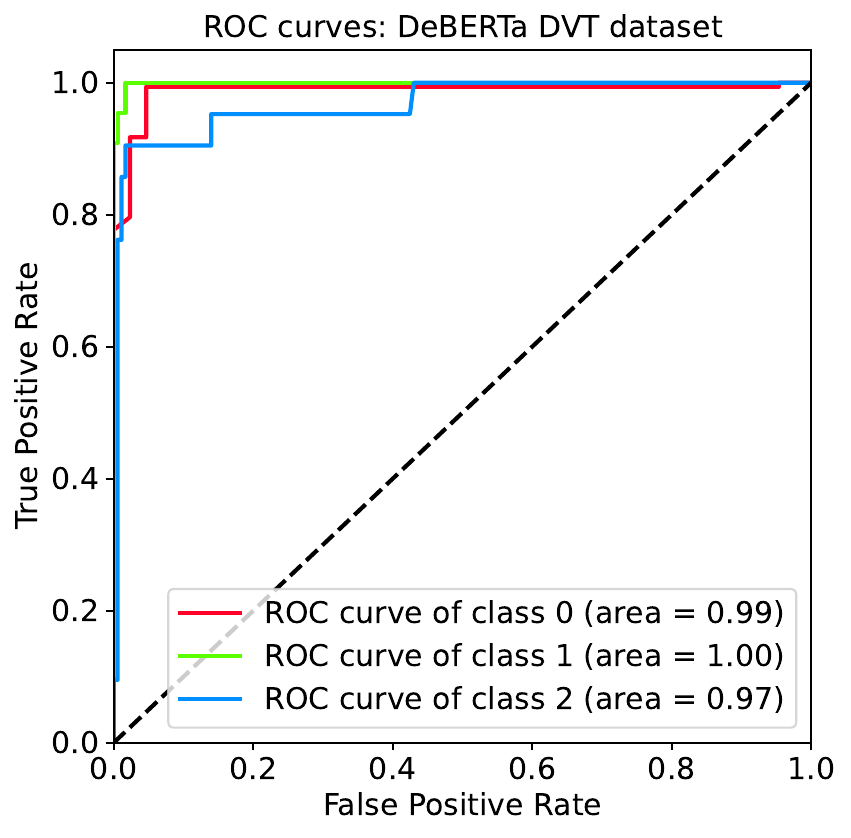}} 
    \end{subfigure}%
    \begin{subfigure}{.33\textwidth}
        \centering
        {\includegraphics[scale=0.38]{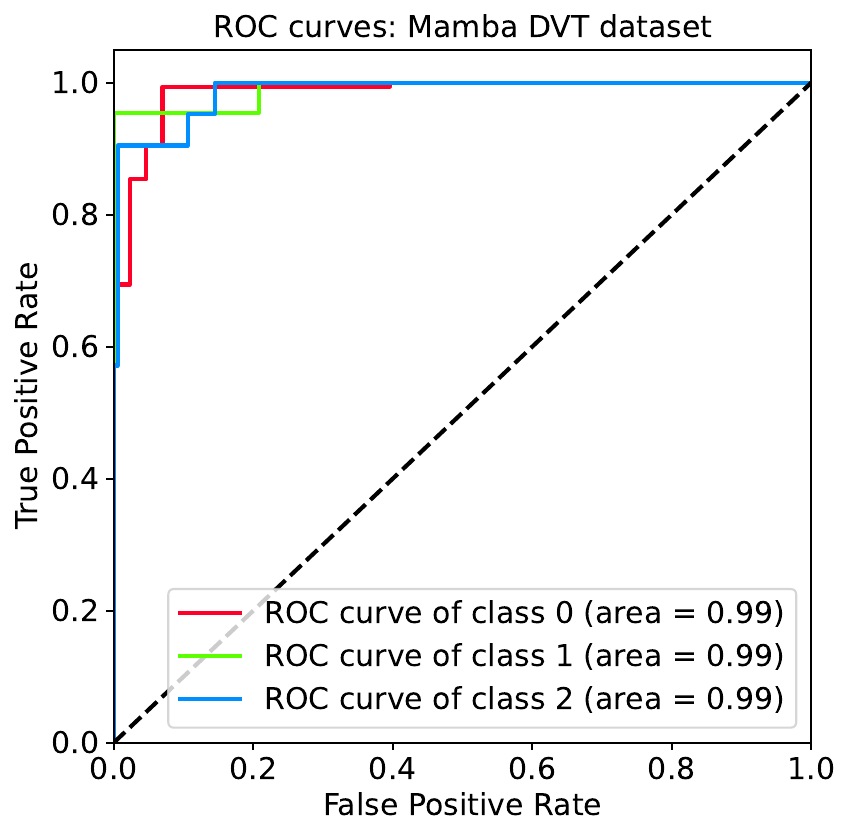}} 
    \end{subfigure}
\caption{ROC curves of different methods on the DVT dataset of Ultrasound reports. (Class 0 - No acute DVT, Class 1 - Upper
extremity acute DVT, and Class 2 - Lower extremity acute
DVT.)}
\label{fig:roc-dvt}
\end{figure*}

\begin{figure*}[ht]
\centering
    \begin{subfigure}{.33\textwidth}
        \centering
        {\includegraphics[scale=0.38]{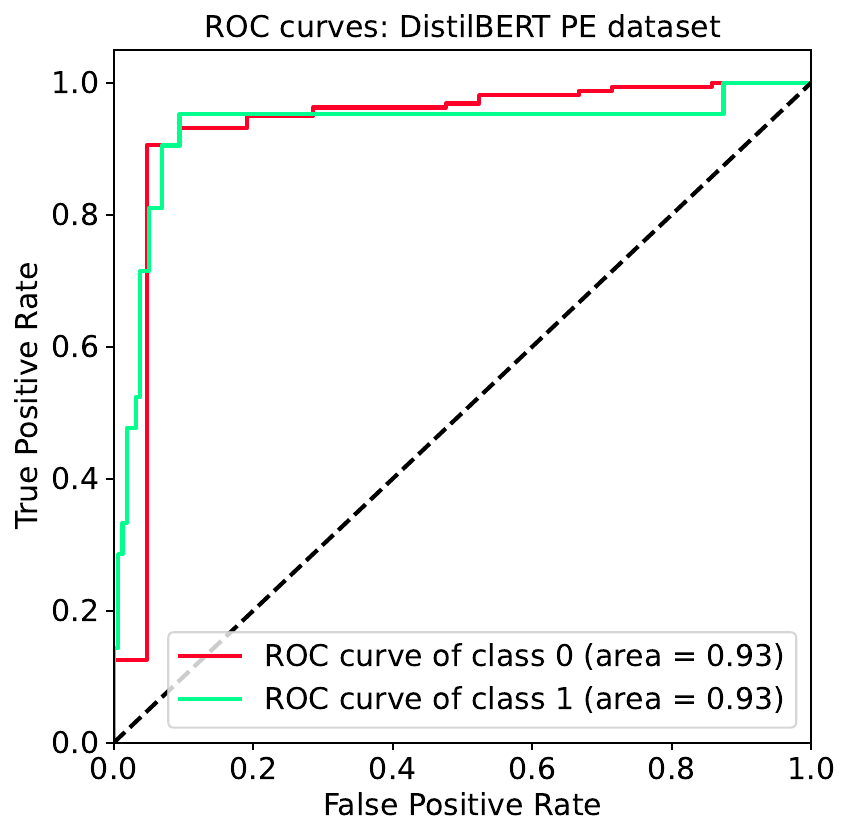}} 
    \end{subfigure}%
    \begin{subfigure}{.33\textwidth}
        \centering
        {\includegraphics[scale=0.38]{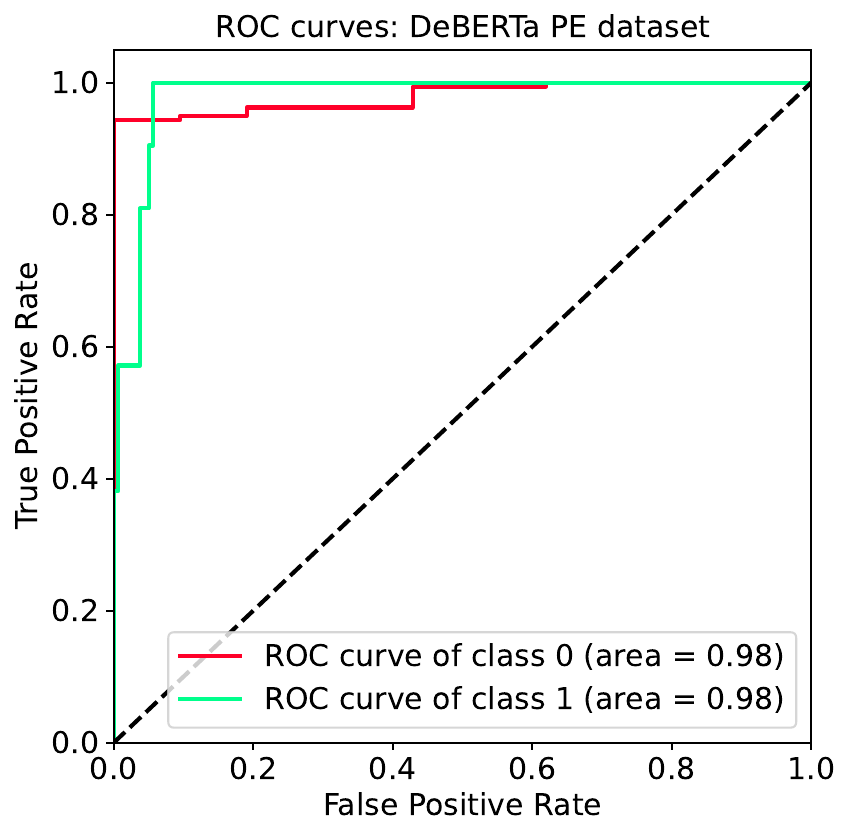}} 
    \end{subfigure}%
    \begin{subfigure}{.33\textwidth}
        \centering
        {\includegraphics[scale=0.38]{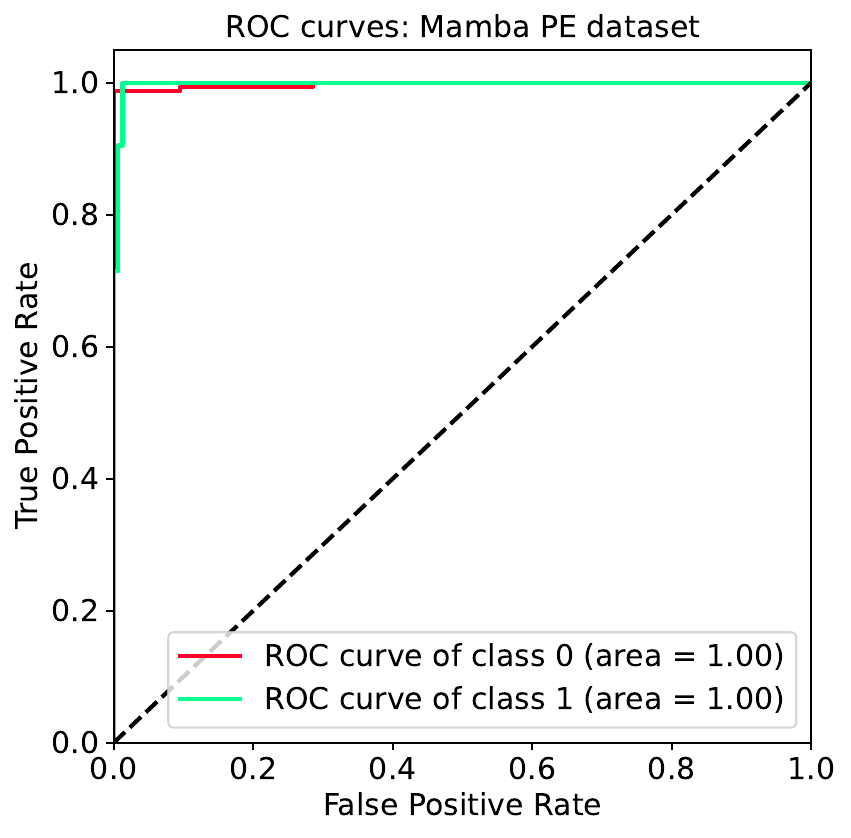}} 
    \end{subfigure}
\caption{ROC curves of different methods on the PE dataset of CT scan reports. (class 0 - No PE, class 1 - PE.)}
\label{fig:roc-pe}
\end{figure*}

The DistilBERT model has 66 million parameters, while the DeBERTa base model has 134 million parameters, comparable to the Mamba-130M model's size. In contrast, the Phi-3 Mini model is significantly larger, designed to handle more complex NLP tasks such as chat-based instructions, beyond just text classification, requiring a more extensive parameter set. The Phi-3 Mini model features a context window of 128K tokens, enabling it to manage long sequence texts effectively.

We perform experiments with aforementioned classifiers and compare the results to those from our previous work, which involved ClinicalBERT and Bi-LSTM model, and a rule-based classifier. 

Table \ref{tab:per-dvt} presents the performance metrics for the experiments conducted on the DVT dataset, including weighted precision, recall, and F1 scores, as well as accuracy, sensitivity, and specificity. The results for the ClinicalBERT and Bi-LSTM models are taken from our previous research. These models have significantly more parameters than Mamba, with a standard BERT model containing 110 million parameters, and a Bi-LSTM consists of even more, depending on the input size and hidden layer dimensions.
The classifiers in the experiment yield comparable results with high accuracy and F1 scores, indicating their effectiveness on the DVT dataset. The dataset, consisting of straightforward and concise medical texts, allows even the smallest DistilBERT model with fewer parameters to achieve results comparable to larger models. During the training phase, the Mamba classifier exhibits signs of overfitting on the training dataset, leading to a slight reduction in sensitivity to 0.92 in the resulting trained model. However, for all other metrics, the Mamba classifier delivers performance on par with the other models. The ClinicalBERT and Bi-LSTM models from our previous research demonstrate the highest sensitivity score but exhibit a slightly lower specificity score.

Figure \ref{fig:roc-dvt} illustrates the ROC curves for the two BERT models and the Mamba model on the DVT dataset. The three models exhibit high performance, with the DistilBERT and DeBERTa models showing a slight bias towards class 1. In contrast, the Mamba model displays more balanced results across all three classes.

In Table \ref{tab:per-pe}, the outcomes of different classifiers on the PE dataset are displayed. Due to the complexity and length of the medical reports in the PE dataset, the BERT models encounter difficulties during training. The input limit of BERT models is 512 tokens, which restricts their ability to process longer texts, leading to truncation during preprocessing. In contrast, the Phi-3 Mini and Mamba models have significantly longer sequence lengths, enabling them to handle the PE reports more effectively than the BERT models. Consequently, language models with extended sequence lengths yield better results in terms of accuracy and F1. The Mamba model demonstrates comparable performance to the hybrid model (combining deep learning and rule-based approaches) from our previous research and exhibits higher sensitivity and specificity.

Figure \ref{fig:roc-pe} presents the ROC curves for the two BERT models and the Mamba model on the PE dataset. The Mamba model significantly outperforms the two Transformer-based models, primarily due to its longer sequence length, which preserves the context of the entire medical reports. In contrast, the DistilBERT model, with its smallest parameter size, performs the worst.

%% file: table1.tex
\begin{table*}[ht]
\centering
\caption{Performance of different techniques on the DVT dataset of Ultrasound reports. }
\begin{tabular}{lccccccr}
\toprule
\textbf{Classifier}   & \textbf{\#Parameters}                                & \textbf{Accuracy} & \textbf{Sensitivity} & \textbf{Specificity} & \textbf{Precision} & \textbf{Recall} & \textbf{F1}    \\ \midrule
DistilBERT     &        66M             & 0.97    & 0.93       & 0.97        & 0.969      & 0.97  & 0.969 \\ 
DeBERTa     &       134M            & 0.975     & 0.93        & 0.978        & 0.976      & 0.975   & 0.975  \\ 
Phi-3 Mini    & 3.8B    &  0.975   &   0.948     &    0.97    & 0.975     &  0.975 &   0.97\\ 
ClinicalBERT + Bi-LSTM \cite{deng2023improving} &  -    & 0.97     & 0.97        & 0.93        & 0.97      & 0.97   & 0.97 \\

 \midrule \textbf{Mamba} & 130M  & 0.97    & 0.92       & 0.965        & 0.97      & 0.97  & 0.969 
  \\ \bottomrule
\end{tabular}
\label{tab:per-dvt}
\end{table*}

%% file: table2.tex
\begin{table*}[ht]
\centering
\caption{Performance of different techniques on the PE dataset of CT Scan reports. }
\begin{tabular}{lccccccr}
\toprule
\textbf{Classifier}     & \textbf{\#Parameters}                               & \textbf{Accuracy} & \textbf{Sensitivity} & \textbf{Specificity} & \textbf{Precision} & \textbf{Recall} & \textbf{F1}    \\ \midrule
DistilBERT          &        66M                & 0.927    & 0.71       & 0.956        & 0.929      & 0.927  & 0.928 \\ 
DeBERTa         &        134M               & 0.938     & 0.76        & 0.96        & 0.94      & 0.938   & 0.939  \\ 
Phi-3 Mini       &        3.8B   &  0.967    &   0.76     &  0.99      & 0.966     & 0.967  & 0.965 \\ 
ClinicalBERT + Bi-LSTM + Rule \cite{deng2023improving} & -  & 0.983 & 0.983 &0.956 &0.984 &0.983 &0.984 \\

 \midrule \textbf{Mamba} &        130M   & 0.98    & 1.0       & 0.97        & 0.98      & 0.98  & 0.978 
  \\ \bottomrule
\end{tabular}
\label{tab:per-pe}
\end{table*}

%% file: related.tex

\section{Related Work}

Conventional NLP systems for text classification relied on either rule-based methods, which required significant manual effort from domain experts for feature selection, or statistical machine learning approaches, which demanded large amounts of training data. Deep learning (DL) has shown promising results in various studies. Many medical text classification tasks have taken advantage of Deep Learning approaches. However, few works have employed DL methods for classifying VTE from medical report datasets, likely due to the limited availability of data. To the best of our knowledge, no previous work has focused on using Mamba models for VTE identification.

\subsection{Traditional approaches} 

Nelson et al. \cite{nelson2015using} integrated statistical machine learning with rule-based NLP methods to detect postoperative VTE in surgical patients at VA hospitals. However, their NLP system was ineffective in accurately identifying postoperative VTE events from clinical notes.  
Sabra et al. \cite{sabra2018prediction} introduced a method called Semantic Extraction and Sentiment Assessment of Risk Factors, which generated feature inputs for a support vector machine classifier aimed at VTE identification. Due to the limited dataset of clinical narratives from electronic health records (EHR), their model achieved an F1 score of only 0.7. 

Shi et al. \cite{shi2021natural} created an NLP system that tokenized patient reports into sentences, identified relevant concepts, and aggregated these semantic representations back to the document and patient level for VTE classification. This approach resulted in an AUC of 0.9 for PE and an AUC of 0.92 for DVT.  
Verma et al. \cite{verma2022developing} utilized an NLP algorithm based on weighted regular expression rules to classify radiologist reports of medical images for VTE, with the rules being manually selected by domain experts. Their methods achieved a PPV of 0.90 and an AUC of 0.96 for DVT identification, while for PE, the results were a PPV of 0.89 and an AUC of 0.96.

\subsection{Deep Learning and Hybrid methods}
Mulyar et al. \cite{mulyar2019phenotyping} investigated various architectures for phenotyping, utilizing BERT representations of free-text clinical notes. Similarly, Olthof et al. \cite{olthof2021machine} found that deep learning-based BERT models outperformed traditional machine learning and rule-based approaches in the classification of radiology reports.  
Goodrum et al. \cite{goodrum2020automatic} extracted text from EHRs and assessed multiple text classification models, including bag-of-words and other machine learning methods. Their findings indicated that a deep learning model using ClinicalBERT yielded the best performance, confirming the effectiveness of deep learning methods in identifying clinically relevant content. 
Lee et al. \cite{lee2019automatic} demonstrated that RNN-based networks were capable of classifying significant findings in radiology reports with high F1 scores.  

In a hybrid study focused on VTE risk factor identification from electronic medical records, Chen et al. \cite{chen2022prediction} employed BERT for word embedding and Bi-LSTM for information extraction, followed by rule-based reasoning to assess PE risk. The experimental results showed that this approach achieved F1 scores of 93.3\% for entity recognition and 94.3\% for relation extraction.  
In our previous work \cite{deng2023improving}, we introduced a deep learning approach for identifying DVT from radiology reports using ClinicalBERT and Bi-LSTM models. Additionally, we proposed a hybrid method combining these deep learning models with a rule-based classifier to detect PE from medical reports. This approach significantly enhanced the accuracy and robustness of PE identification, especially in dealing with imbalanced datasets, resulting in high F1 scores.

\subsection{Mamba language models}
Grazzi et al. \cite{grazzi2024mamba} utilized Mamba models for both simple function estimation and natural language processing tasks that involve learning from context. Their findings demonstrated that the Mamba models' performance matches that of Transformer networks in these applications. 
Yang et al. \cite{yang2024clinicalmamba} leverage the linear computational efficiency of Mamba models to handle long sequences of clinical notes, reaching up to 16K tokens. They pre-trained the model on the MIMIC-III \cite{johnson2016mimic} dataset, and evaluating its capabilities in cohort selection and ICD coding tasks. The results highlighted that the Clinical Mamba model outperforms both the standard Mamba and the clinical Llama models, particularly when dealing with longer sequences of clinical text.
Lu et al. \cite{lu2023efficient} explore the application of SSM in the classification of long texts, tackling the efficiency issues associated with the Transformer architecture. They showed that their SSM approach matches the performance of attention based models while being approximately 36\% more efficient. Furthermore, their method proved to be robust against noisy inputs, even under severe conditions. Song et al. \cite{song2024state} demonstrated that Mamba based model outperformed BERT models in long text classification tasks while also achieving higher efficiency.

%% file: future.tex
\section{Future Works}
In order to address the challenges of efficiently deploying Language Models in clinical environments, while our current study highlights Mamba's effectiveness in classifying VTE, the practical application of such models in real-world healthcare settings requires further refinement. One key area of focus should be the optimization of these models for deployment across various clinical settings, including both resource-constrained edge devices and cloud-based systems used in hospital settings.

To achieve this, future research could explore techniques such as model pruning and quantization. These methods can significantly reduce the model's size and computational demands, making it more suitable for deployment on devices with limited resources. For example, pruning could help remove non-essential components of the model, while quantization reduces memory usage by lowering the precision of the model's weights. By applying these methods, a model like Mamba, which originally requires 520 MB of memory with 32-bit precision, could see its memory footprint reduced to 130 MB through the use of 8-bit quantization. This would represent a 75\% reduction in memory usage, making it far more efficient and practical for deployment in a wide range of clinical settings.

Another important area for future research is knowledge distillation, where a smaller model is trained to replicate the outputs of a larger, more complex model. This process allows for the creation of lightweight models that maintain high performance while being less resource-intensive. Such models are especially suitable for real-time clinical applications, where computational efficiency is crucial.


%% file: conclusion.tex
\section{Conclusion}
In this study, we employ the Mamba architecture-based classifier to effectively identify VTE based on free-text clinical reports from medical imaging. Additionally, we evaluate the performance of one lightweight LLMs (Phi-3 Mini) in classifying VTE, which also delivers comparable results. This research builds upon our previous work, which utilized a hybrid approach involving deep learning (ClinicalBERT and Bi-LSTM models) and a rule-based classifier to identify PE from medical reports. While the hybrid method is effective in terms of performance, it had a complex architecture that involves multiple components. Particularly, the rule-based classifier requires careful manual feature selection by domain experts, which limits its generalization ability when applying the method to other medical domains. In contrast, the new Mamba architecture offers an efficient and effective approach to both training and inference on complex and lengthy texts.

The Mamba model achieved impressive results, with 97\% accuracy and F1 score on the DVT dataset and 98\% accuracy and F1 score on the PE dataset. It delivers comparable results to the hybrid method while eliminating the need for hand-engineered rules, thereby reducing the model's complexity. The Phi-3 Mini classifier also outperforms the two BERT models. However, this LLM has a significantly larger number of parameters, making it computationally intensive. For classification tasks involving long texts, LLM is not an efficient method. The experimental findings support the effectiveness of Mamba-based models for VTE identification.